\documentclass{article}
\usepackage{ijcai26}

\usepackage{times}
\usepackage{soul}
\usepackage{url}
\usepackage[hidelinks]{hyperref}
\usepackage[utf8]{inputenc}
\usepackage[small]{caption}
\usepackage{graphicx}
\usepackage{amsmath}
\usepackage{amsthm}
\usepackage{booktabs}
\usepackage[switch]{lineno}
\usepackage[utf8]{inputenc} 
\usepackage[T1]{fontenc}    
\usepackage{hyperref}       
\usepackage{url}            
\usepackage{booktabs}       
\usepackage{amsfonts}       
\usepackage{nicefrac}       
\usepackage{microtype}      
\usepackage{xcolor}         
\usepackage{xspace}
\usepackage{graphicx}
\usepackage{amsmath}
\usepackage{amssymb}
\usepackage{array}
\usepackage{wrapfig}
\usepackage{subfigure}
\usepackage{pifont}
\usepackage{multirow}
\usepackage{enumitem}
\usepackage{color, colortbl}
\definecolor{Gray}{gray}{0.9}

\usepackage{algorithm}
\usepackage{algpseudocode}  

\definecolor{fgreen}{RGB}{177,207,149}
\definecolor{fred}{RGB}{234,179,138}

\definecolor{firstcolor}{RGB}{20,128,85}
\definecolor{secondcolor}{RGB}{20,104,168}
\definecolor{thirdcolor}{RGB}{236,84,20}

\newcommand{\ourmethod}{FedCIGAR\xspace}

\title{FedCIGAR: A Personalized Reconstruction Approach \\ for Federated Graph-level Anomaly Detection}

\author{
Yunfeng Zhao$^{1*}$
\and
Yixin Liu$^{2*}$
\and
Qingfeng Chen$^{1\dagger}$
\and
Shiyuan Li$^2$\and
Yue Tan$^2$\And
Shirui Pan$^2$\\
\affiliations
$^1$Guangxi University\\
$^2$Griffith University\\
\emails
\{yunf.zhao, qingfeng\}@gxu.edu.cn,\\
\{yixin.liu, shiyuan.li, yue.tan, s.pan\}@griffith.edu.au
}

\begin{document}

\maketitle
\renewcommand{\thefootnote}{\fnsymbol{footnote}}
\footnotetext[1]{All authors contributed equally to this research.}
\footnotetext[2]{Corresponding author.}
\begin{abstract}
Graph-level anomaly detection (GLAD) is crucial for ensuring the reliability of graph-driven applications by identifying abnormal graphs that deviate from the majority. Considering the privacy concerns in distributed scenarios, federated graph-level anomaly detection (FedGLAD) has emerged as a promising solution to enable collaborative detection without sharing raw data. However, existing methods suffer from poor generalization due to the reliance on unrealistic synthetic anomalies and insufficient personalization capabilities under data heterogeneity. To address these challenges, we propose a novel \textbf{Fed}erated graph-level anomaly detection approach with \textbf{C}luster-adapt\textbf{I}ve \textbf{GA}ted \textbf{R}econstruction (\ourmethod). Specifically, we design a reconstruction-based paradigm trained on normal graphs to avoid synthetic data. Furthermore, we introduce a client-side node contribution gating mechanism and a server-side sliding window-based clustering strategy to tackle data heterogeneity. Extensive experiments demonstrate that \ourmethod achieves superior performance and robustness in contrast to state-of-the-art methods. 
\end{abstract}

\section{Introduction}
Graph-structured data have been widely used in a variety of real-world scenarios, such as bioinformatics, small molecules, and social networks~\cite{stokes2020deep,barabasi2004network,easley2010networks}. To ensure the reliability and robustness of graph-driven applications, graph-level anomaly detection (GLAD) is a crucial task on graph data~\cite{zhao2023using,luo2022deep,ijcai2022p305}, which aims to identify the abnormal graphs that are different from the majority. Recently, most GLAD methods~\cite{kim2024rethinking,zhang2022dual,ma2023towards,liu2023towards} follow a centralized training principle, which requires collecting all graph data for model training. However, in real-world scenarios, graph data are usually distributed among different users. Owing to privacy protection concerns~\cite{sarasamma2005hierarchical}, users are often unwilling to share their data, which greatly hinders the application of centralized GLAD methods in practical scenarios.

\begin{figure} [t]
    \centering
    \subfigure[Synthetic abnormal]{\label{subfig:intro_1}
        \includegraphics[height=3cm]{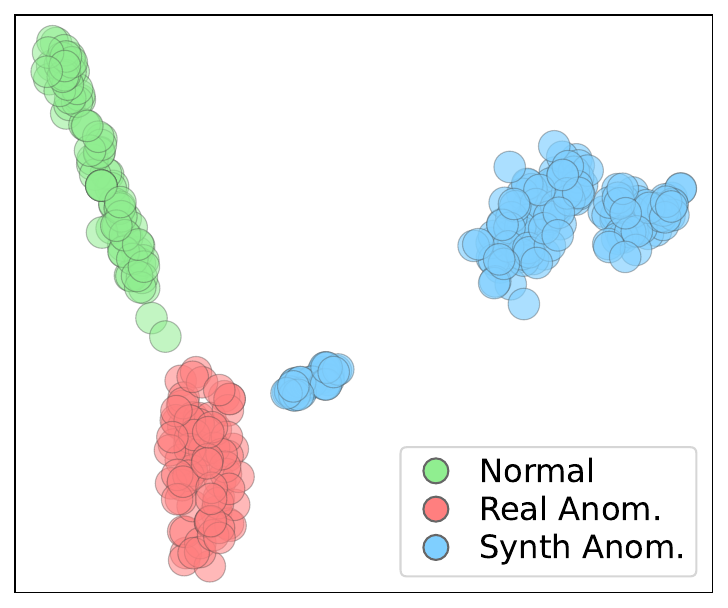}
    }\hfill
    \subfigure[Client data distributions]{\label{subfig:intro_2}
        \includegraphics[height=3cm]{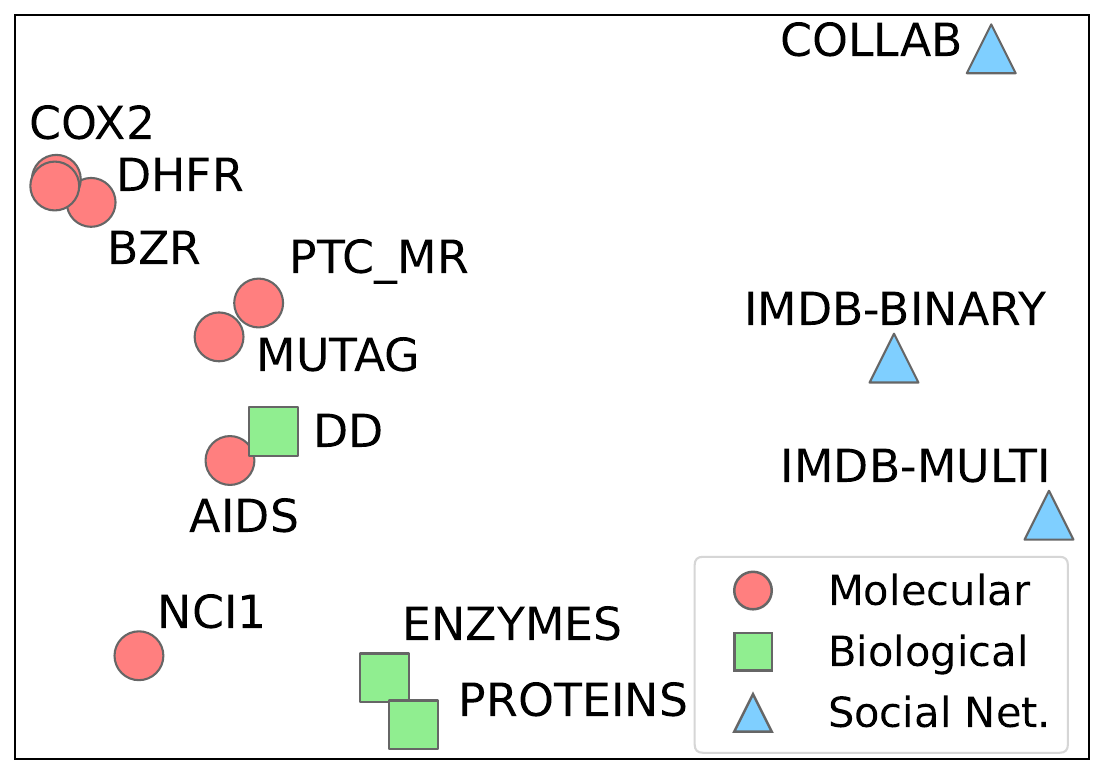}
    }
    \caption{(a): T-SNE visualization of normal graphs, synthetic anomalies, and real anomalies learned by LG-FGAD on AIDS dataset. (b): Heterogeneous client data distributions across different datasets.}
    \label{fig:intro}
\end{figure}

As an effective distributed learning paradigm, federated learning (FL)~\cite{ijcai2023p426,fu2022federated,li2021survey} provides a promising pathway to address the above limitation. Recently, an emerging branch of studies has attempted to incorporate FL and GLAD to mitigate the impact of centralized training and facilitate anomaly detection in distributed environments. These federated graph-level anomaly detection (FedGLAD) methods perform federated optimization on specially designed anomaly detection models to enable effective GLAD under distributed data settings. 
For instance, FGAD~\cite{cai2024towards} employs a teacher-student architecture with knowledge distillation to maintain the personalization of local models and enable collaborative learning among clients. To improve anomaly detection in the absence of annotated anomalous graphs, it uses self-generated anomalies to simulate realistic anomalous cases for model training. LG-FGAD~\cite{cai2024lg} indicates that node-level representations may induce potential anomalies and proposes a local-global anomaly awareness module, which captures anomalous patterns at both node and graph levels to enhance the discrimination ability.

Despite their effectiveness, the existing FedGLAD methods treat anomaly detection as a binary classification task and employ artificially generated anomalies to mimic abnormal samples, which leads to their \textit{\textbf{Limitation 1} - poor generalization capacity to realistic and heterogeneous anomaly cases}. Considering the sparsity of anomalous graph data, current GLAD methods~\cite{ma2022deep,liu2023towards} usually adopt a one-class training setting~\cite{yin2024mcm}, where abnormal samples are unavailable for model training. Under this constraint, to train the binary classification-based detection models without annotated samples, the existing FedGLAD methods use an additional generative model to synthesize pseudo-anomalous samples. Nevertheless, these synthetic anomalies may fail to reflect the true distribution of real-world abnormal graphs, which hinders their ability to capture diverse and realistic anomaly characteristics. As visualized in Fig.~\ref{subfig:intro_1}, learned by LG-FGAD~\cite{cai2024lg}, the representations of synthetic anomalies significantly deviate from those of real anomalies, indicating the inherent limitation of such a data synthesis paradigm.

\renewcommand{\thefootnote}{\arabic{footnote}}
\footnotetext[1]{Code is available at \url{https://github.com/yunf-zhao/FedCIGAR}}

In addition to the limited distribution modeling capability of local GLAD models, the FedAvg-style global aggregation paradigm in existing FedGLAD methods further intensifies \textit{\textbf{Limitation 2} – insufficient personalization capability under heterogeneous client data}. As demonstrated in Fig.~\ref{subfig:intro_2}, client data in real-world scenarios usually exhibits heterogeneous distributions (see Appendix A for details), which makes it challenging for one global model to capture all diverse anomaly patterns~\cite{xie2021federated}. However, current FedGLAD methods adopt a FedAvg-like paradigm~\cite{mcmahan2017communication} that globally updates a single server model for all clients, which results in limited adaptability to heterogeneous client data. Even if the knowledge distillation-based collaborative learning mechanism in FGAD and LG-FGAD can preserve the personalization to some extent, the globally shared student model may still fail to adapt well to the diverse anomaly distributions across clients.

To tackle the aforementioned limitations, in this paper, we propose \textbf{\ourmethod}, a \textbf{Fed}erated graph-level anomaly detection approach with \textbf{C}luster-adapt\textbf{I}ve \textbf{GA}ted \textbf{R}econstruction. To address \textit{\textbf{Limitation 1}}, we design a reconstruction-based GLAD model that can be directly trained on normal graph data without relying on synthetic anomalies. This formulation eliminates the need for costly data generation and avoids the adverse impact of low-quality pseudo-anomalous samples. To better capture structural and attributive anomaly patterns, we propose a \textit{feature–structure joint reconstruction model} for GLAD that independently encodes structural and feature information into a unified latent space and then reconstructs the original features and graph structures accordingly. 
To handle \textit{\textbf{Limitation 2}}, we enhance the personalization capability of both the client-side detection model and the server-side aggregation mechanism. At the client side, we introduce a \textit{node contribution gating mechanism} that equips each client with an independent local model to assess the contribution of different nodes to anomaly scores. This mechanism alleviates the limitation of global reconstruction models under heterogeneous client data and preserves client-specific anomaly characteristics. At the server side, we introduce a \textit{sliding window–based clustering aggregation strategy} that dynamically groups clients with similar data distributions and aggregates their updates in a cluster-adaptive manner. In this way, \ourmethod effectively balances global collaboration and local personalization under heterogeneous client data. 
In summary, this paper makes the following contributions:
\begin{itemize}
    \item \textbf{New FedGLAD Paradigm.} We, for the first time, introduce a reconstruction-based paradigm for FedGLAD. Compared to classification-based methods, our approach avoids synthetic anomaly generation and achieves better generalization to different anomaly patterns.
    \item \textbf{Data Heterogeneity Handling.} We tackle the data heterogeneity across clients from both server and client perspectives by crafting a node contribution-based model and a cluster-adaptive server aggregation mechanism.
    \item \textbf{Superior Detection Performance.} We conduct extensive experiments on multiple datasets and multi-dataset settings, where \ourmethod achieves excellent detection performance and robustness.
\end{itemize}

\section{Preliminaries}
\noindent \textbf{Notations.}
Let $\mathcal{D}=\{D_1, \dots,D_C \}$ represent the federated datasets distributed in $C$ clients, where $\mathcal{D}_c=\{\mathcal{G}_1, \dots,\mathcal{G}_{N_c} \}$ represents the local dataset held by client $c$ and contains $N_c$ graphs. Each graph $\mathcal{G}_i\in\mathcal{D}$ has a node set $\mathcal{V}_i$ and an edge set $\mathcal{E}_i$. The connection between nodes of graph $\mathcal{G}_i$ can be represented by an adjacency matrix $\mathbf{A}_i\in \{0,1\}^{n_i \times n_i}$, where $n_i=|\mathcal{V}_i|$ is the nodes number in graph $\mathcal{G}_i$. The features matrix of graph $\mathcal{G}_i$ can be described by $\mathbf{X}_i \in \mathbb{R}^{n_i \times m}$ where the $j$-th row $\mathbf{x}_{i,j}\in \mathbb{R}^d$ represents the feature vector for node $v_j\in \mathcal{V}_i$.

\noindent \textbf{Problem Formulation.} This paper concentrates on the one-class (a.k.a. unsupervised) federated graph-level anomaly detection (FedGLAD) problem. Given a federated dataset $\mathcal{D}$, where each client possesses its own graph set $\mathcal{D}_c$, and it can be divided into anomalous graph set $\mathcal{D}_a$ and normal graph set $\mathcal{D}_n$, with $\mathcal{D}_c=\mathcal{D}_a \cup \mathcal{D}_n$, $\mathcal{D}_a \cap \mathcal{D}_n = \emptyset$, and $|\mathcal{D}_a|=N_a \ll |\mathcal{D}_n|=N_n$. Formally, the objective of FedGLAD is to learn a set of scoring functions $f_{\mathcal{W}}=\{f_{\mathcal{W}^{(1)}}, \dots,f_{\mathcal{W}^{(C)}} \}$ (where $\mathcal{W}^{(c)}$ is the set of learnable parameters of client $c$), which assign anomaly scores to indicate the abnormality of graphs from each client.
The overall training objective can be regarded as minimizing the following loss function: 
\begin{equation} \label{eq:problem}
\begin{aligned}
    \min_{\mathcal{W}^{(1)}, \ldots, \mathcal{W}^{(C)}} 
    \frac{1}{C} \sum_{c=1}^{C} 
    \frac{|D_{c}|}{|D|}
    \bigl(
            \ell_{c}(y_n, f_{\mathcal{W}^{(c)}}(\mathcal{G}_n)) \\
            + \ell_{c}(y_a, f_{\mathcal{W}^{(c)}}(\mathcal{G}_a))
    \bigr),
\end{aligned}
\end{equation}

\noindent where $\{\mathcal{G}_n;y_n\}$ represents the normal graph labeled with $y$=1, and $\{\mathcal{G}_a;y_a\}$ represents the anomalous graph labeled with $y$=0. Here, $f_{\mathcal{W}^{(c)}}( \cdot )$ and $\ell_{c}( \cdot )$ separately denote the $c$-$th$ client scoring function and loss function.

\section{Methodology}
\begin{figure*}
\vspace{-3mm}
    \centering
        \includegraphics[width=.94\linewidth]{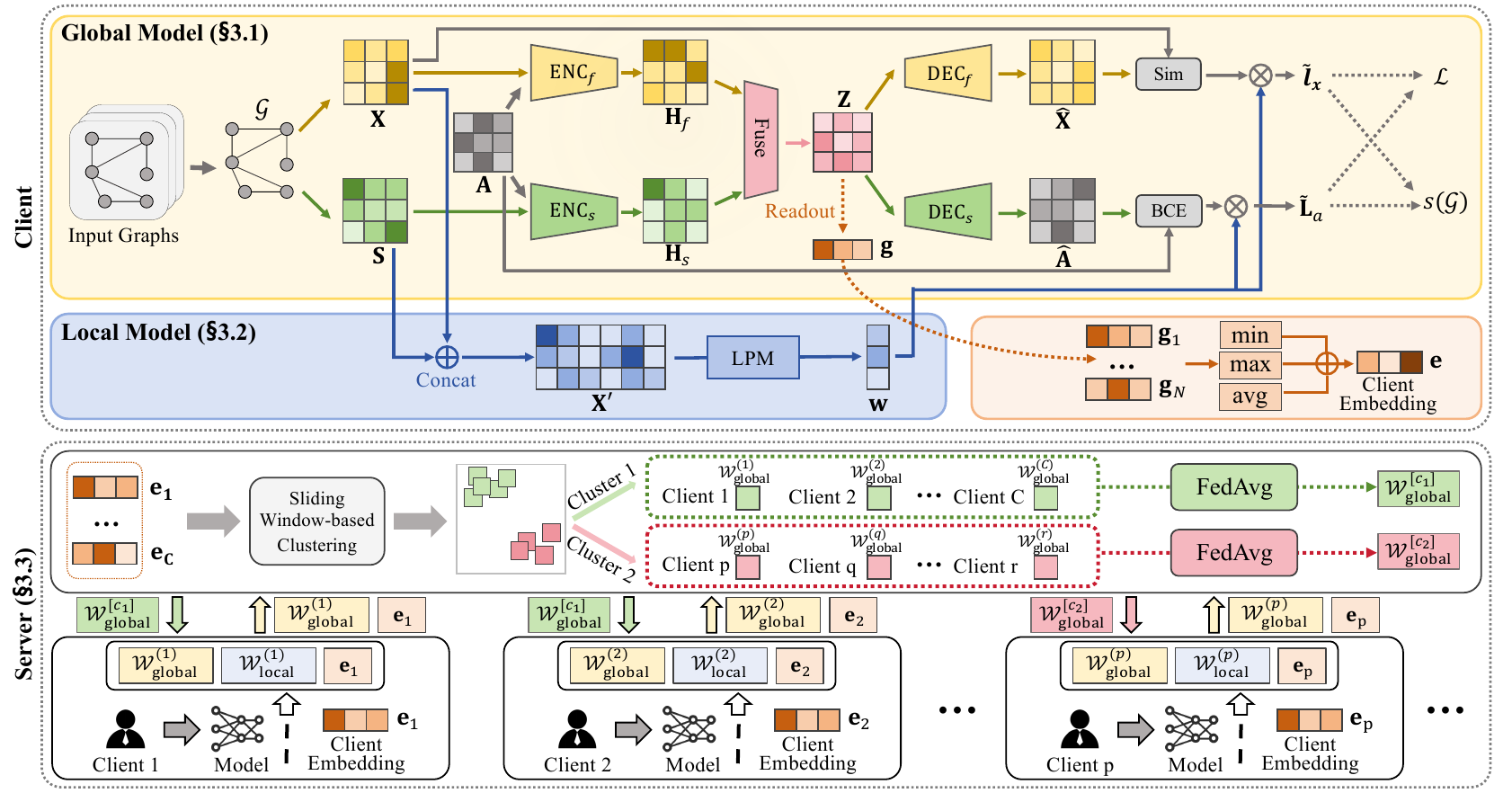}
        \vspace{-1mm}
    \caption{The overall pipeline of \ourmethod.}
	\label{fig:pipeline}
    \vspace{-3mm}
\end{figure*}

In this section, we introduce \ourmethod, a novel reconstruction-based method for federated graph-level anomaly detection (FedGLAD). The overall pipeline of \ourmethod is illustrated in Fig.~\ref{fig:pipeline}. To capture diverse anomaly patterns without relying on real anomalous samples, we introduce a \textit{feature-structure joint reconstruction model}~(Sec.~\ref{subsec:reconstruction}), which explicitly captures the attributive and structural information of the graph and reconstructs them from a fused latent space. Once well trained, the reconstruction error can serve as an effective indicator of graph abnormality. 
To alleviate the impact of data heterogeneity across clients, we design a \textit{node contribution gating module}~(Sec.~\ref{subsec:gate}) to gauge each node's contribution to graph anomaly under personalization, which allows each client to maintain adaptive anomaly scoring tailored to its local data distribution. 
Furthermore, we introduce a \textit{sliding window-based clustering aggregation strategy}~(Sec.~\ref{subsec:cluster}) at the server side, which adaptively groups clients with similar behaviors and performs cluster-wise aggregation to better accommodate heterogeneous client data.

\subsection{Feature-Structure Joint Reconstruction}\label{subsec:reconstruction}

In order to train the binary classification-based federated anomaly detection models using only normal data, the existing FedGLAD methods (e.g., FGAD~\cite{cai2024towards} and LG-FGAD~\cite{cai2024lg}) adopt data generation techniques to synthesize pseudo-anomalous samples for supervision. However, such generated anomalies usually fail to reflect real-world anomaly patterns, leading to suboptimal detection performance. 
To address the limitations caused by relying on synthetic anomalies, \ourmethod abandons the binary classification paradigm and instead employs a reconstruction-based FedGLAD model. 
The core idea of our detection model is to separate normal and anomalous graph patterns based on the reconstruction assumption, i.e., the reconstruction model trained on purely normal data can accurately rebuild normal graphs while producing large reconstruction errors for anomalous graphs~\cite{kim2024rethinking}. In order to effectively capture both structural and attributive patterns of graphs, we carefully design a feature–structure joint reconstruction model for GLAD, which is detailed in the following paragraphs. For simplicity, we only describe the reconstruction process of an individual graph~$\mathcal{G}$ and omit its index (e.g.,~$i$) in the notation.

\noindent\textbf{Dual-Channel Encoder.}
The existing reconstruction models for GLAD~\cite{kim2024rethinking,liu2023towards} typically use GNN-based autoencoder to aggregate \textit{raw node features} for graph reconstruction. However, the low-dimensional features of the real-world graph datasets (e.g., atom types) may provide limited semantic information. In some specific domains (e.g., social networks), node features are even absent, which provides limited informative resources for modeling data patterns. To mitigate the impact of poor-quality features and explicitly capture structural patterns, we introduce structure encoding as an additional input of our reconstruction model. Specifically, for a graph $\mathcal{G}$, its structure encoding matrix is denoted as $\mathbf{{S}} \in \mathbb{R}^{n \times d_s} = \operatorname{concat}[\mathbf{S}^{\operatorname{DSE}}, \mathbf{{S}}^{\operatorname{RWSE}}]$, 
\noindent where each row $\mathbf{s}_{j}$ is the concatenation of $\mathbf{s}_{j}^{\operatorname{DSE}}$, the one-hot encoding of  node $v_j$'s degree, and $\mathbf{s}_{j}^{\operatorname{RWSE}}$, the random walk diffusion-based structure encoding~\cite{dwivedigraph} of node $v_j$. Taking the structural encoding as an auxiliary input not only enhances the informativeness of the original features but also injects structural knowledge from both local (i.e., node degree) and global (i.e., random-walk characteristics) perspectives.

After obtaining the structural encoding, we employ two parallel channels to learn feature and structure representations: 

\begin{equation}\label{eq:encoder}
\mathbf{H}_f = \operatorname{ENC}_f(\mathbf{X}, \mathbf{A}), \quad
\mathbf{H}_s = \operatorname{ENC}_s(\mathbf{S}, \mathbf{A}),
\end{equation}

\noindent where $\operatorname{ENC}_f(\cdot)$ and $\operatorname{ENC}_s(\cdot)$ denote the GNN-based feature encoder and structure encoder, respectively, while $\mathbf{H}_f$ and $\mathbf{H}_s$ respectively denote the feature and structure representations of graph $\mathcal{G}$. Such a decoupled design allows the model to learn complementary representations, capturing attribute semantics and structural patterns independently.

\noindent\textbf{Fused Decoder.}
After obtaining the feature and structure representations, we fuse them into a shared latent space, where the complementary information from both domains is integrated to form a unified representation for downstream reconstruction, i.e., $\mathbf{Z} = \operatorname{MLP}\Big(\operatorname{concat}\big[\mathbf{H}_f, \mathbf{H}_s\big]\Big)$.  
Taking $\mathbf{Z}$ as input, we use a GNN-based feature decoder $\operatorname{DEC}_f(\cdot)$ and a dot-product-based structure decoder $\operatorname{DEC}_s(\cdot)$ to reconstruct the raw feature matrix and adjacency matrix, respectively:
\begin{equation}\label{eq:decoder}
\hat{\mathbf{X}} = \operatorname{DEC}_f(\mathbf{Z}), \quad
\hat{\mathbf{A}} = \operatorname{DEC}_s(\mathbf{Z}),
\end{equation}
where $\hat{\mathbf{X}}$ denotes the reconstructed feature matrix and $\hat{\mathbf{A}}$ denotes the reconstructed adjacency matrix.

\noindent\textbf{Statistics-Aware Reconstruction.} 
Although minimizing the average reconstruction loss over all nodes in a graph can facilitate model optimization, recent studies~\cite{kim2024rethinking} pinpoint that such a strategy may suffer from the ``reconstruction flip'' phenomenon, i.e., the model yields even lower reconstruction errors in particular nodes on unseen test graphs than on training graphs. As a result, the reconstruction error may fail to faithfully indicate the abnormality of test graphs under federated settings. 
To bridge the gap, we introduce a statistics-aware reconstruction strategy for our GLAD model, which not only incorporates the mean reconstruction loss but also considers the standard deviation of reconstruction errors. For model training, the loss function can be written by: 

\begin{equation}
    \mathbf{l}_x = \left[ 1 - \frac{\mathbf{X}_{i,:}^\top \hat{\mathbf{X}}_{i,:}}{\|\mathbf{X}_{i,:}\|_2 \cdot \|\hat{\mathbf{X}}_{i,:}\|_2}\right]_{i\in [|\mathcal{V}|]} \in \mathbb{R}^{|\mathcal{V}|},
\label{eq:Lx}
\end{equation}
\begin{equation}
\resizebox{.91\linewidth}{!}{$\displaystyle
    \mathbf{L}_a = \left[\mathbf{A}_{i,j} \log \hat{\mathbf{A}}_{i,j} + (1 - \mathbf{A}_{i,j}) \log (1 - \hat{\mathbf{A}}_{i,j})\right]_{i,j\in [|\mathcal{V}|]} \in \mathbb{R}^{|\mathcal{V}|^{2}}
$},
\label{eq:La}
\end{equation}
\begin{equation}                                                                                                                    
\resizebox{.91\linewidth}{!}{$\displaystyle
    \mathcal{L} = \alpha \big( \operatorname{avg}(\mathbf{l}_x) + \beta\, \operatorname{std}(\mathbf{l}_x) \big) + (1-\alpha)\big( \operatorname{avg}(\mathbf{L}_a) + \beta\, \operatorname{std}(\mathbf{L}_a) \big)
$},
\label{eq:loss}
\end{equation}

\noindent where $|| \cdot ||$ represents the $l2$ norm, $\hat{\mathbf{X}}_{i,:}$ stands for $i$-th node's reconstruction feature and $\hat{\mathbf{A}}_{i,j}$ represents the reconstructed adjacency entry between node $i$ and $j$, while $\alpha$ and $\beta$ are the balance hyperparameters. When testing, the anomaly score $s$ of graph $\mathcal{G}$ can be directly calculated by the mean and standard deviation of the reconstruction error, i.e., $s(\mathcal{G})=\mathcal{L}$.

Through additionally modeling the standard deviation of reconstruction errors, the model can encourage stable and uniform reconstruction errors on all nodes in a graph, thereby alleviating the reconstruction flip issue by suppressing excessively high or low reconstruction errors. Meanwhile, variance modeling enables \ourmethod to better capture the distributional characteristics of reconstruction errors, which leads to more reliable anomaly characterization.

\subsection{Node Contribution Gating}\label{subsec:gate}

By collaboratively optimizing a global anomaly detection model in a federated manner without sharing raw data, clients can benefit from collective knowledge while preserving data privacy. Although such collaborative learning can train a global GLAD model to capture broader data patterns, it may overlook client-specific anomaly characteristics under heterogeneous data distributions. On one client, anomalies may predominantly stem from a small subset of structurally influential nodes (e.g., popular users in social networks), whereas on another client, anomalies may arise from nodes with specific features (e.g., certain types of atoms in molecular graphs). In this case, a global reconstruction model may be insufficient to capture the heterogeneous anomaly behaviors exhibited by different clients, leading to sub-optimal performance.

To address this issue, we introduce a node contribution gating module as the personalized model for each client. Specifically, a local personalized model is employed to estimate node-wise contribution weights conditioned on the local data distribution. Unlike the reconstruction model that is globally shared, the parameters of the local personalized model are maintained independently on each client, enabling client-specific adaptation to heterogeneous anomaly patterns. 
For each client, the input of the local personalized model is constructed by concatenating the feature and structure encoding matrices, i.e., $\mathbf{X}' = \operatorname{concat}(\mathbf{X}, \mathbf{S})$. Afterwards, the GNN-based local personalized model $\operatorname{LPM}(\cdot)$ is built to estimate the contribution of each node within its local context:

\begin{equation}\label{eq:last}
\mathbf{w} = \operatorname{sigmoid}\!\left( {\operatorname{LPM}\!\big(\mathbf{X}',\mathbf{A}\big)} / {\tau} \right) \in \mathbb{R}^{n},
\end{equation}

\noindent where $\mathbf{w}$ denotes the contribution weight vector (in which each element indicates the contribution of a corresponding node) and $\tau$ is the temperature hyperparameter to adjust the smoothness of the node-wise weights. 

Given the column weight vector $\mathbf{w}$, the reconstruction errors in Eqs.~\eqref{eq:Lx} and \eqref{eq:La} can be further reweighted as: 

\begin{equation}
\tilde{\mathbf{l}}_x = \mathbf{l}_x \odot \mathbf{w}, \quad
\tilde{\mathbf{L}}_a = \mathbf{L}_a \odot (\mathbf{w}\mathbf{w}^T),
\end{equation}

\noindent where $\odot$ indicates element-wise product. The reweighted errors, i.e., $\tilde{\mathbf{l}}_x$ and $\tilde{\mathbf{L}}_a$, can be further used to calculate the loss function and anomaly score, replacing the original reconstruction errors ${\mathbf{l}}_x$ and ${\mathbf{L}}_a$. 

According to such a gating mechanism, the local model can learn to adaptively weight node contribution according to the local data patterns. For instance, the atoms that are closely associated with abnormal behaviors in a client with molecular data can be assigned higher weights. In this manner, the model realizes personalized anomaly detection on each client while continuing to benefit from the powerful global knowledge learned by the GLAD model at server.

\subsection{Sliding Window-based Clustering Aggregation}\label{subsec:cluster}

While the above gating mechanism can alleviate data heterogeneity via introducing personalized local models, the client-level heterogeneity across different data distributions can still persist if we rely on a single global reconstruction model. For instance, the reconstruction process for molecular graphs can be significantly different from that for social network graphs, and using one unified global model may fail to adapt to such fundamentally diverse data characteristics. 
A promising solution to this issue is clustered FL, which groups clients with similar data distributions and performs cluster-wise model aggregation. 
However, existing clustered FL algorithms typically rely on high-dimensional variables for communication, such as model parameters~\cite{wang2024fedsg} or gradients~\cite{xie2021federated}, and thus often converge slowly due to their top-down bi-partitioning mechanisms, which inevitably incur high communication costs. Although FLT~\cite{jamali2022federated} acknowledged the above issue, its static clustering strategy, which performs clustering only once at model initialization, may fail to capture changes in dynamic cluster relations during multi-round communication.

To address the above issues, we propose a sliding window-based clustering aggregation strategy for FedGLAD. Our strategy consists of two steps: client representation construction and dynamic clustering. In the first step, we introduce a low-dimensional and high-density client representation to compactly characterize clients’ behavioral patterns and reduce communication overhead. Then, according to these representations, we dynamically cluster clients with similar characteristics for aggregation.

\noindent\textbf{Client Representation Construction.}  To describe the characteristics of each client, we compute statistics over its graph representations to construct a client representation. Based on the node-level representations $\mathbf{Z}$ learned from the reconstruction model, which capture both feature and structure information, we apply a readout function to obtain the graph-level representation $\mathbf{g}$ for graph $\mathcal{G}$. Further, for each client $c$, we collect the graph representations of all training samples, denoted as:
$\mathbb{G}_c=\{\mathbf{g}_i \mid i=1,\dots,N_c\}$. 
Then, the embedding of client $c$ is calculated by statistically aggregating its graph representations:

\begin{equation}\label{eq:client_embed}
\mathbf{e}_c = \left[
\operatorname{mean}(\mathbb{G}_c) \| 
\operatorname{max}(\mathbb{G}_c) \| 
\operatorname{min}(\mathbb{G}_c)
\right], 
\end{equation}

\noindent where $\mathbf{e}_c$ is the embedding of client $c$. 

\noindent\textbf{Dynamic Clustering-based Aggregation.} 
After obtaining the low-dimensional client representations, we design the server aggregation strategy based on them. Compared to binary clustering~\cite{xie2021federated}, k-means optimizes a global partition across all clients, exhibiting greater robustness and yielding more stable client groupings~\cite{ghosh2020efficient}. Accordingly, we adopt k-means clustering to partition clients.  Specifically, at communication round $t$, the server performs k-means on the stacked client representations $\mathbf{E} = [\mathbf{e}_1; \ldots; \mathbf{e}_C]$ to obtain the clustering assignment $\mathcal{A}_t = \{\mathcal{C}_1,\dots,\mathcal{C}_K\}$ where $K$ denotes the number of clusters. Unfortunately, k-means still requires periodic updates to accommodate evolving client representations, incurring considerable computational costs on the server side.
\noindent To mitigate re-clustering issue, we design a two-phase clustering module with update and silent phases. The server maintains a sliding window $\mathcal{Q} = {\mathcal{A}_1,\dots,\mathcal{A}_L}$ to track historical clusterings, where $L$ is the maximum length. At communication round $t$, the server first establishes whether $\mathcal{Q}$ is full. If not, the module enters the update phase: the server computes the current clustering $\mathcal{A}_t$ via k-means and evaluates clustering stability by calculating the minimum Adjusted Rand Index (ARI) with historical clusterings:

\begin{equation}\label{eq:ari}
R_t = \min_{\mathcal{A}_i \in \mathcal{Q}} ARI(\mathcal{A}_t, \mathcal{A}_i).
\end{equation}

\noindent If $R_t < \theta$, where $\theta$ is a predefined stability threshold, the current clustering deviates substantially from historical patterns, and the server clears $\mathcal{Q}$ and re-initializes it with $\mathcal{A}_t$. Otherwise, the server appends $\mathcal{A}_t$ to $\mathcal{Q}$. Once $\mathcal{Q}$ is full, the module enters the silent phase: the server clears $\mathcal{Q}$ and directly reuses the final stable clustering $\mathcal{A}_L$ for the subsequent $P$ rounds without executing k-means, thereby substantially reducing server-side computational overhead while preserving stable client groupings. The overall training and testing algorithms and complexity analysis are provided in Appendices B.1 and B.2, respectively.

\section{Experiments}
\begin{table*}[ht]
\centering
\resizebox{\textwidth}{!}{
\begin{tabular}{@{}c|cc|cc|cc|cc@{}}
\toprule
\multirow{2}{*}{\textbf{Methods}} 
& \multicolumn{2}{c|}{\textbf{IMDB-BINARY}}
& \multicolumn{2}{c|}{\textbf{MUTAG}}
& \multicolumn{2}{c|}{\textbf{DD}}
& \multicolumn{2}{c}{\textbf{AIDS}} \\
\cline{2-9}
\addlinespace[0.7ex]
& AUC & F1-Score
& AUC & F1-Score
& AUC & F1-Score
& AUC & F1-Score \\
\midrule
Self-train
& $65.01\pm1.53$ & $60.83\pm1.55$
& $94.44\pm2.08$ & $89.58\pm3.61$
& $80.11\pm0.12$ & $74.52\pm0.31$
& $98.35\pm0.16$ & $97.89\pm0.0$ \\
\midrule
FedAvg
& $40.96 \pm 3.44$ & $45.95 \pm 1.66$
& $85.83 \pm 2.36$ & $85.00 \pm 0.00$
& $30.83 \pm 0.44$ & $34.16 \pm 1.52$
& $61.26 \pm 0.54$ & $58.80 \pm 0.60$ \\
FedProx
& $41.86 \pm 0.32$ & $45.70 \pm 1.96$
& $84.17 \pm 2.36$ & $85.00 \pm 0.00$
& $30.81 \pm 0.43$ & $33.91 \pm 1.45$
& $61.18 \pm 0.37$ & $58.80 \pm 0.60$ \\
\midrule
GCFL
& $56.98 \pm 5.56$ & $45.44 \pm 2.25$
& $83.33 \pm 3.11$ & $78.33 \pm 4.71$
& $38.56 \pm 0.13$ & $30.06 \pm 1.10$
& $74.31 \pm 0.91$ & $70.86 \pm 1.76$ \\
FedStar
& $54.76 \pm 1.28$ & $52.40 \pm 1.04$
& $85.42 \pm 10.82$ & $83.33 \pm 10.27$
& $55.88 \pm 0.92$ & $55.95 \pm 1.98$
& $86.75 \pm 2.11$ & $82.77 \pm 1.23$ \\
\midrule
FGAD
& $67.90\pm7.74$ & $65.08\pm5.27$
& $83.75\pm3.64$ & $80.24\pm4.68$
& $81.03\pm0.89$ & $75.33\pm0.98$
& $99.25\pm0.11$ & $98.14\pm0.62$ \\
LG-FGAD
& $66.40 \pm 3.55$ & $63.10 \pm 3.15$
& $87.50 \pm 4.37$ & $85.00 \pm 0.00$
& $81.66 \pm 0.16$ & $75.72 \pm 0.27$
& $\mathbf{99.57 \pm 0.06}$ & $97.55 \pm 1.34$ \\
\midrule
FedCIGAR
& $\mathbf{70.25 \pm 1.01}$ & $\mathbf{65.16 \pm 1.07}$
& $\mathbf{97.52 \pm 1.71}$ & $\mathbf{91.67 \pm 0.00}$
& $\mathbf{83.13 \pm 0.79}$ & $\mathbf{76.79 \pm 0.51}$
& $99.33 \pm 0.09$ & $\mathbf{98.84 \pm 0.33}$ \\
\bottomrule
\end{tabular}
}
\caption{Anomaly detection performance in terms of AUC and F1-Score (in percent, mean$\pm$std) under the single-dataset setting.}
\label{tab:main_results_single}
\end{table*}

\subsection{Experimental Setup}
\noindent\textbf{Datasets.} 
Following the setting in Cai et al.~\shortcite{cai2024lg}, we validate the performance of our method on two types of datasets. (1) The single-dataset setting indicates IID scenarios, where a single dataset is randomly distributed across multiple clients, each of which holds a subset of the dataset. We employ four graph classification datasets, including AIDS, DD, MUTAG, and IMDB-BINARY. (2) The multi-dataset setting indicates non-IID scenarios, where multiple datasets consist of several single datasets. Each client holds a specific graph dataset. We employ four multi-dataset collections, including MOLECULES, BIOCHEM, SMALL, and Mix. The first class in each dataset is treated as anomalous, and more dataset details are illustrated in Appendix C.1. 

\noindent\textbf{Baselines.}
We compare \ourmethod with seven baselines including (1) Self-train, where each client trains locally without any communications; (2) FedAvg~\cite{mcmahan2017communication} and (3) FedProx~\cite{li2020federated}, the widely used federated learning methods; (4) GCFL~\cite{xie2021federated} and (5) FedStar~\cite{tan2023federated}, two federated graph learning methods; (6) FGAD~\cite{cai2024towards} and (7) LG-FGAD~\cite{cai2024lg}, two state-of-the-art FedGLAD methods. Part of the results are borrowed from \cite{cai2024towards,cai2024lg}. For datasets not included in these papers, we use random search to select the hyperparameters.

\noindent\textbf{Evaluation Metrics and Implementation.} 
Following the existing FedGLAD method~\cite{cai2024lg}, we consider two evaluation metrics: Area Under the Curve (AUC) and F1-Score. We report the average score across 10 trials. More implementation details are given in Appendix C.2.

\begin{table*}[t]
\centering
\resizebox{\textwidth}{!}{
\begin{tabular}{@{}c|cc|cc|cc|cc@{}}
\toprule
\multirow{2}{*}{\textbf{Methods}} 
& \multicolumn{2}{c|}{\textbf{MOLECULES}}
& \multicolumn{2}{c|}{\textbf{BIOCHEM}}
& \multicolumn{2}{c|}{\textbf{SMALL}}
& \multicolumn{2}{c}{\textbf{MIX}} \\
\cline{2-9}
\addlinespace[0.7ex]
& AUC & F1-Score
& AUC & F1-Score
& AUC & F1-Score
& AUC & F1-Score \\
\midrule

Self-train 
& $70.35\pm0.56$ & $66.38\pm0.68$ 
& $68.3\pm0.74$ & $61.67\pm1.31 $
& $57.92\pm4.56$ & $51.12\pm4.11$ 
& $59.8\pm0.33$ & $52.14\pm0.32$ \\
\midrule

FedAvg
& $54.41 \pm 3.21$ & $55.57 \pm 1.46$
& $47.49 \pm 1.04$ & $51.24 \pm 1.42$
& $48.90 \pm 0.60$ & $52.77 \pm 0.64$
& $47.96 \pm 0.61$ & $53.89 \pm 0.83$ \\

FedProx 
& $57.93 \pm 2.14$ & $55.91 \pm 0.44$
& $46.04 \pm 0.49$ & $51.71 \pm 1.59$
& $48.89 \pm 0.50$ & $52.42 \pm 0.27$
& $46.79 \pm 0.63$ & $53.50 \pm 0.88$ \\
\midrule

GCFL 
& $58.86 \pm 1.09$ & $57.80 \pm 1.20$
& $51.44 \pm 1.18$ & $54.88 \pm 1.67$
& $53.93 \pm 0.51$ & $57.68 \pm 0.02$
& $51.46 \pm 0.96$ & $55.35 \pm 1.14$ \\

FedStar
& $57.03 \pm 2.02$ & $55.54 \pm 0.86$
& $47.80 \pm 0.48$ & $53.21 \pm 0.84$
& $51.09 \pm 2.00$ & $55.28 \pm 1.92$
& $51.68 \pm 1.61$ & $54.89 \pm 1.11$ \\
\midrule

FGAD
& $62.00 \pm 5.13$ & $60.00 \pm 4.78$ 
& $61.41 \pm 1.78$ & $55.20 \pm 1.77$ 
& $62.38 \pm 1.98$ & $56.27 \pm 2.23$ 
& $59.58 \pm 0.31$ & $52.46 \pm 0.59$ \\

LG-FGAD
& $70.84 \pm 0.47$ & $65.72 \pm 0.59$
& $67.98 \pm 0.16$ & $61.52 \pm 0.81$
& $66.39 \pm 0.91$ & $\mathbf{62.22 \pm 1.30}$
& $62.26 \pm 0.97$ & $\mathbf{56.52 \pm 0.39}$ \\
\midrule

FedCIGAR
& $\mathbf{74.22 \pm 1.37}$ & $\mathbf{69.52 \pm 1.46}$
& $\mathbf{72.20 \pm 1.95}$ & $\mathbf{65.03 \pm 1.40}$
& $\mathbf{70.03 \pm 1.21}$ & $61.11 \pm 1.12$
& $\mathbf{63.77 \pm 2.04}$ & $54.96 \pm 1.68$ \\
\bottomrule

\end{tabular}
}
\caption{Anomaly detection performance in terms of AUC and F1-Score (in percent, mean$\pm$std) under the multi-dataset setting.}
\label{tab:main_results_multi}
\end{table*}

\subsection{Performance Comparison}

The comparison results on four single-dataset and four multi-dataset are reported in Table~\ref{tab:main_results_single} and Table~\ref{tab:main_results_multi}. From these results, we have the following observations. \ding{182}~\ourmethod demonstrates superior performance compared to state-of-the-art approaches across four single-dataset, with the sole exception of a slight reduction in AUC on the AIDS dataset(0.24\%). In most cases, it achieves better results for both AUC and F1-Score compared to the best baseline (e.g., an increase of $\uparrow$10.02\% in AUC and $\uparrow$6.67\% in F1-Score on the MUTAG dataset). These results demonstrate the effectiveness of our framework in the single-dataset setting. \ding{183} In the multi-dataset scenarios, client data from various domains, where non-IID problem is further aggravating, \ourmethod remains significantly superior to other methods. For instance, on BIOCHEM, \ourmethod outperforms the strongest baseline, LG-FGAD, by 4.22\% in AUC and 3.51\% in F1-Score. These results suggest that \ourmethod effectively captures shared information across cross-domain clients, enhancing detection performance.

\begin{figure} [tbp]
    \centering
    \subfigure[MUTAG\label{subfig:weight_mutag}]{
        \includegraphics[scale=0.28]{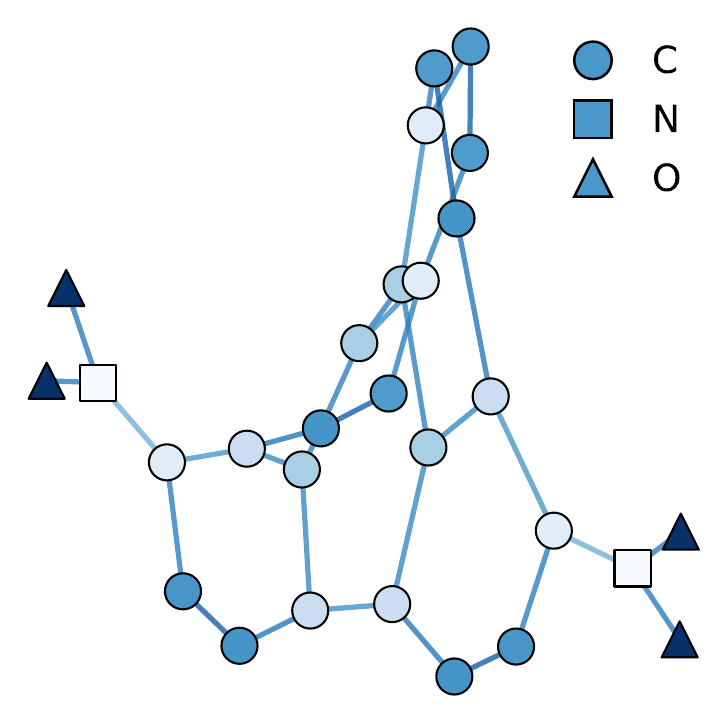}
    }\hfill
    \subfigure[PTC\_MR\label{subfig:weight_ptc}]{
        \includegraphics[scale=0.28]{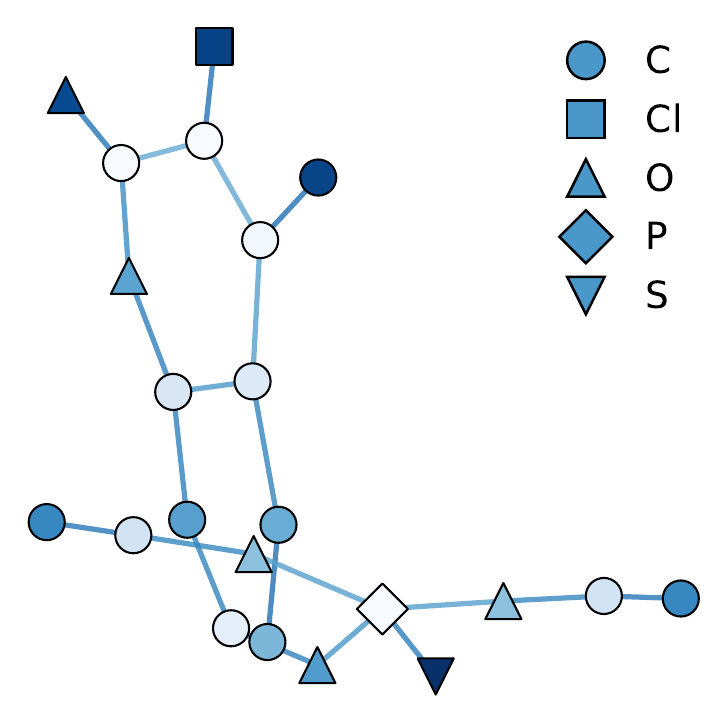}
    }\hfill
    \caption{Visualization of node weight.}
    \label{fig:visual_weight}
\end{figure}
\begin{table}[t]
\centering
\small
\setlength{\tabcolsep}{6pt}
\resizebox{\columnwidth}{!}{
\begin{tabular}{@{}c|cccc@{}}
\toprule
\textbf{Method} 
& \textbf{MOLECULES} 
& \textbf{BIOCHEM}
& \textbf{MIX}
& \textbf{MUTAG}  \\
\midrule
\ourmethod 
& $\mathbf{74.22 \pm 1.37}$
& $\mathbf{72.20 \pm 1.95}$
& $\mathbf{63.77 \pm 2.04}$
& $\mathbf{97.52 \pm 1.71}$ \\
\midrule
w/o Struct 
& $70.70 \pm 1.06$
& $68.16 \pm 3.24$
& $60.45 \pm 1.94$
& $93.98 \pm 0.65$ \\
w/o Gate 
& $63.32 \pm 1.36$
& $62.86 \pm 3.52$
& $58.87 \pm 0.56$
& $96.76 \pm 0.66$ \\
w/o Cluster 
& $73.83 \pm 1.28$
& $71.13 \pm 1.29$
& $61.85 \pm 1.92$
& $97.38 \pm 0.66$ \\
\bottomrule
\end{tabular}
}
\caption{Performance of \ourmethod and variants in terms of AUC.}
\label{tab:ablation}
\end{table}
\subsection{Ablation Study}
To verify the effectiveness of each component in FedCIGAR, we conduct an ablation study to compare \ourmethod with three variants: \ding{182}~\textbf{w/o SE:} removing the structure encoding branch; \ding{183}~\textbf{w/o Gating:} removing the anomaly gate module; and \ding{184}~~\textbf{w/o Cluster:} replacing the sliding clustering module with the FedAvg parameter aggregation method.

The experimental results are shown in Table~\ref{tab:ablation}. We can observe that all components are effective for the overall performance. \ding{182}~Removing the structure encoding  (\textbf{w/o SE}) leads to significant performance degradation, demonstrating that high-quality node representations are crucial for effectively distinguishing anomalies. \ding{183}~Removing the gating module (\textbf{w/o Gate}) results in a substantial performance drop across multi-datasets, which confirms the importance of capturing the client's specific patterns. \ding{184}~The single-dataset shows a light decrease compared to the multi-dataset after removing the sliding clustering mechanism (\textbf{w/o Cluster}), which confirms that the naive FedAvg algorithm fails in non-IID scenarios.

\subsection{Visualization Analysis}

\noindent\textbf{Node Contribution.} To investigate the weight allocation mechanism of the gating module in \ourmethod, we visualize the node contribution weights on the MUTAG and PTC\_MR datasets from SMALL. As shown in Fig.~\ref{subfig:weight_mutag}, \ourmethod tends to assign extreme weights to N and O atoms, and assign uniform weights for C atoms. This is consistent with the patterns of MUTAG dataset, in which the ``NO$_2$'' functional group is a key structural cue for distinguishing mutagenic molecules. 
Fig.~\ref{subfig:weight_ptc} shows the weight allocation for PTC\_MR dataset, where \ourmethod primarily detects anomalies through special atoms (e.g., Cl, O, and S) rather than the carbon chain. In summary, the gating module enables \ourmethod to adaptively capture node pattern most relevant to anomalies.

\begin{figure}[tbp]
    \centering    \subfigure[MOLECULES\label{subfig:cluster_molecules}]{
        \includegraphics[scale=0.23]{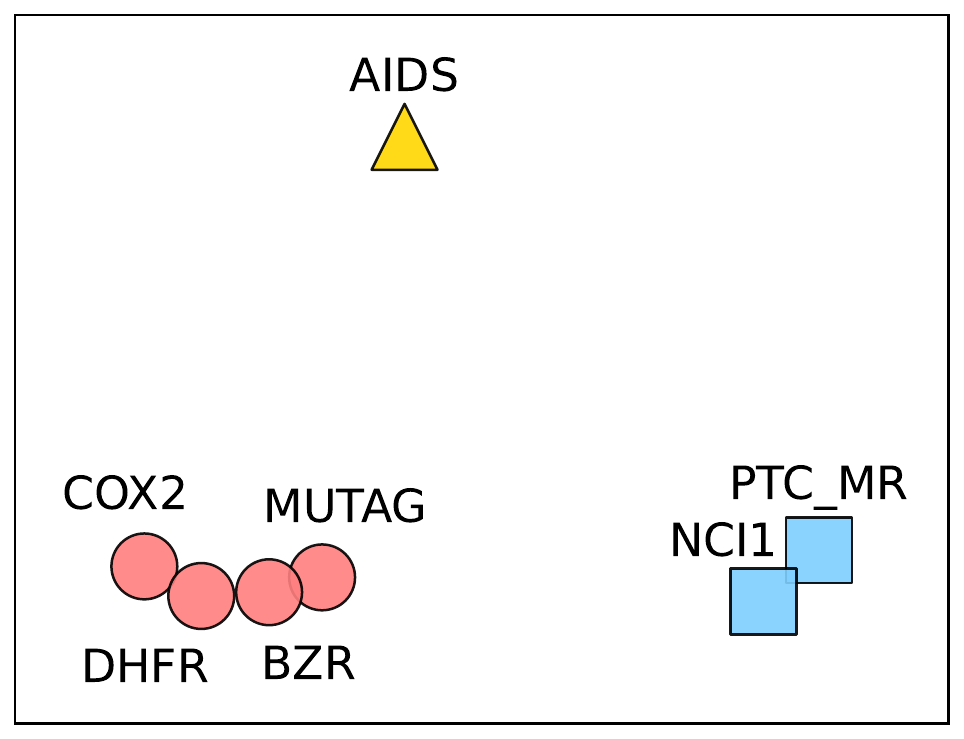}
    }\hfill    \subfigure[MIX\label{subfig:cluster_mix}]{
        \includegraphics[scale=0.23]{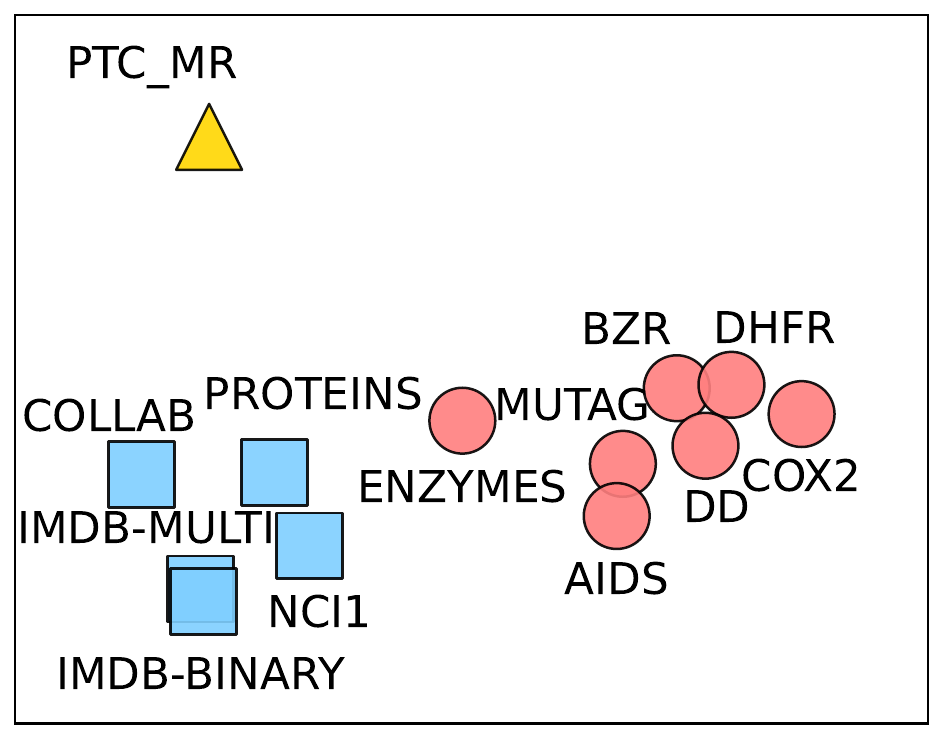}
    }\hfill
    \caption{Visualization of Cluster.}
    \label{fig:visual_cluster}
\end{figure}
\noindent\textbf{Cluster.} To better understand the clustering mechanism of \ourmethod, we visualize the client embedding distributions on both MOLECULES and MIX datasets in Fig.~\ref{fig:visual_cluster}. We can witness that \ourmethod adaptively partitions clients into multiple homogeneous clusters, where compact intra-cluster embeddings and clear inter-cluster separation. These results indicates that the clustering mechanism can explicitly distinguish homogeneous and heterogeneous clients based on their representations, enabling effective aggregation of shared anomaly patterns across datasets and even domains, thereby improving local anomaly detection performance.

\subsection{Parameter Analysis}
\begin{figure} [tbp]
\vspace{-6mm}
    \centering
    \subfigure[SMALL\label{subfig:hp_small}]{
        \includegraphics[scale=0.23]{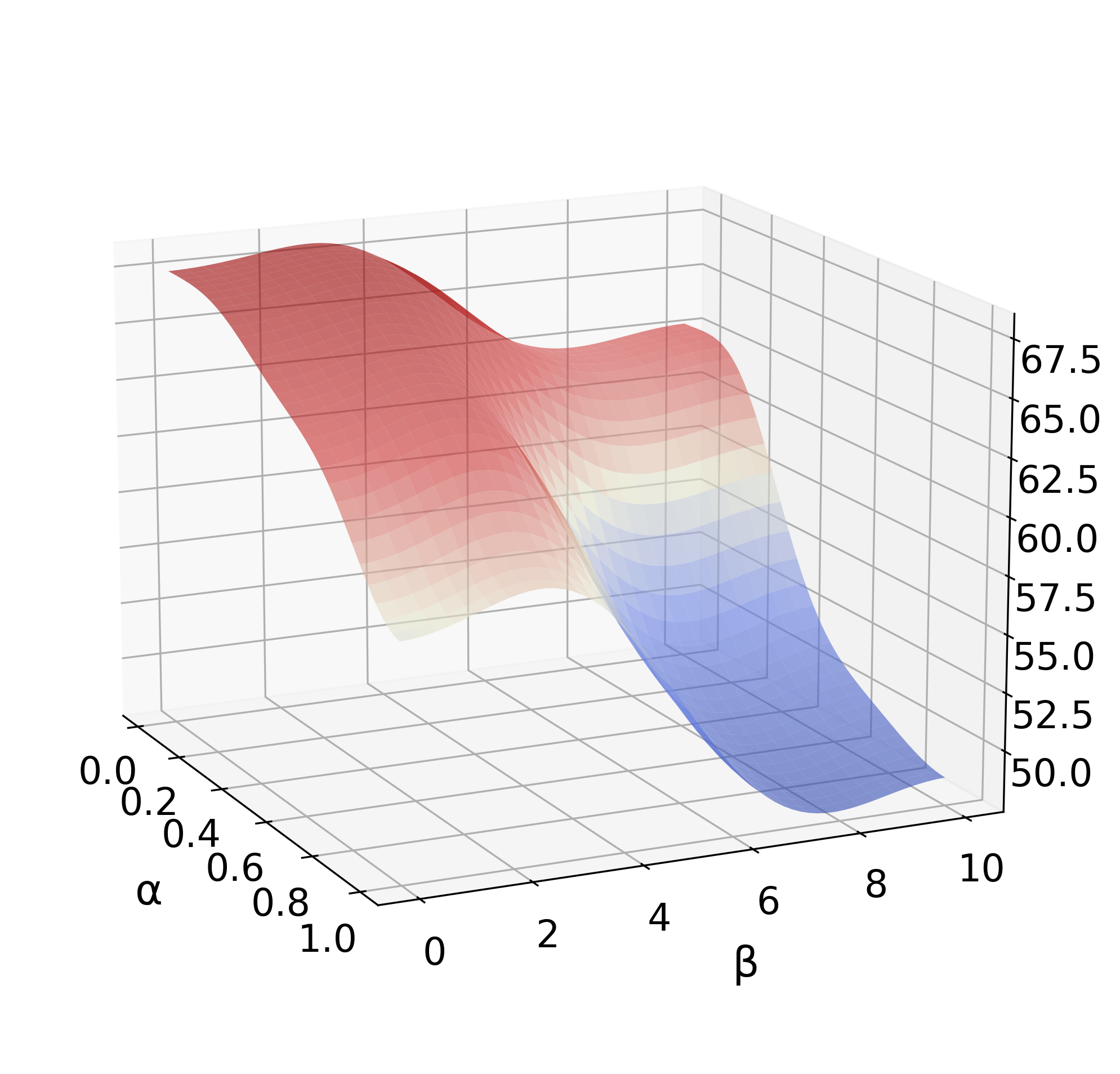}
    }\hfill
    \subfigure[DD\label{subfig:hp_DD}]{
        \includegraphics[scale=0.23]{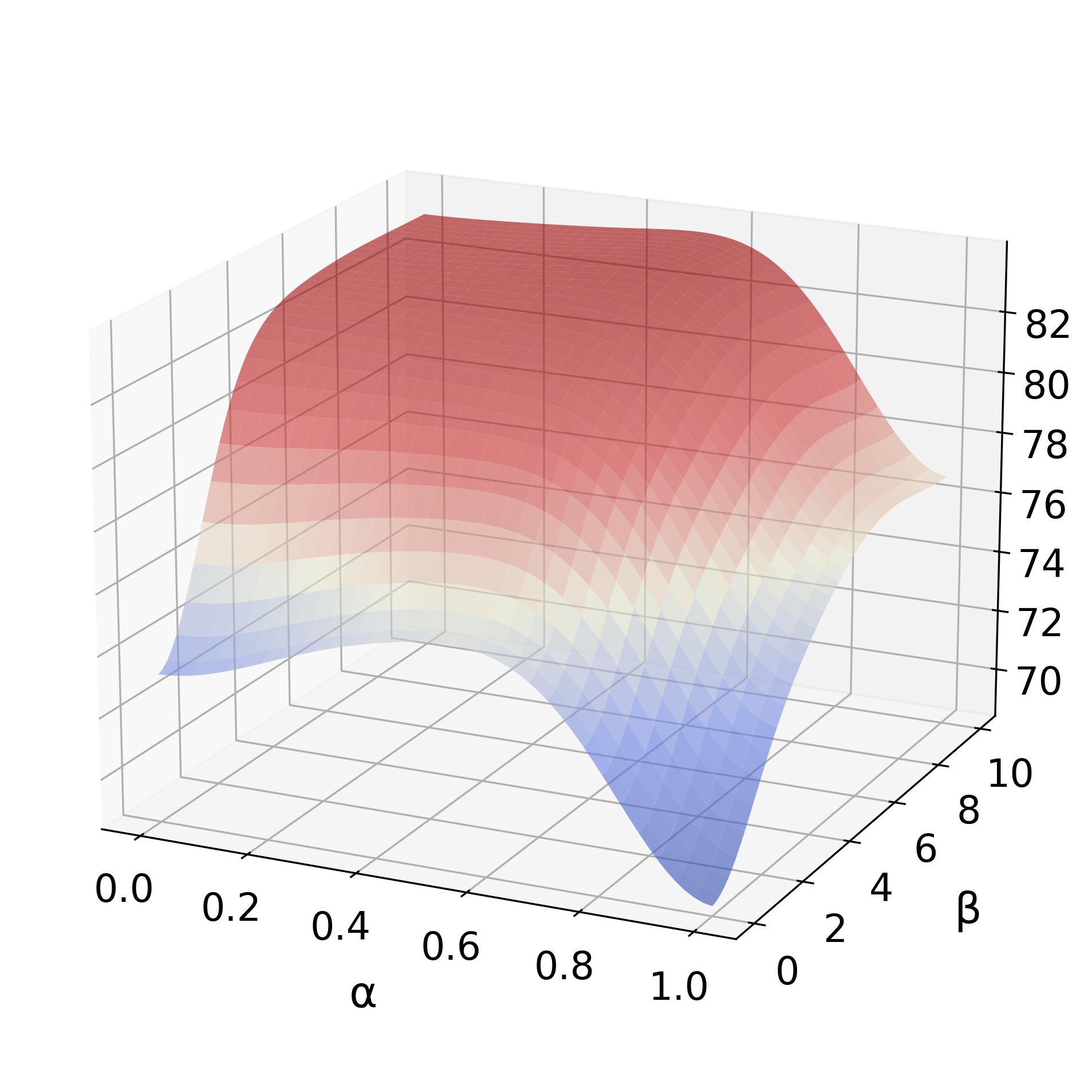}
    }\hfill
    \vspace{-3mm}
    \caption{The sensitivity of \ourmethod in terms of $\alpha$ and $\beta$.}
    \label{fig:hyperparams}
\end{figure}
\noindent\textbf{Balance Hyperparameters $\alpha$ and $\beta$.} Fig.~\ref{fig:hyperparams} presents the sensitivity of $\alpha$ and $\beta$ on SMALL and DD datasets, showing clear dataset-dependent trends. SMALL favors smaller $\alpha$ and $\beta$, indicating the preference for structural information and reconstruction mean in anomaly detection, whereas DD prefers higher $\alpha$ and $\beta$ values, relying more on feature information and reconstruction variance. These divergences reveal pronounced heterogeneity in anomaly patterns across graph datasets, and \ourmethod's multi-aspect modeling improves performance while reducing reliance on any single metric.
\begin{figure} [t]
    \centering
    \subfigure[IMDB-BINARY\label{subfig:ClientNum_IMDB-BINARY}]{
        \includegraphics[scale=0.27]{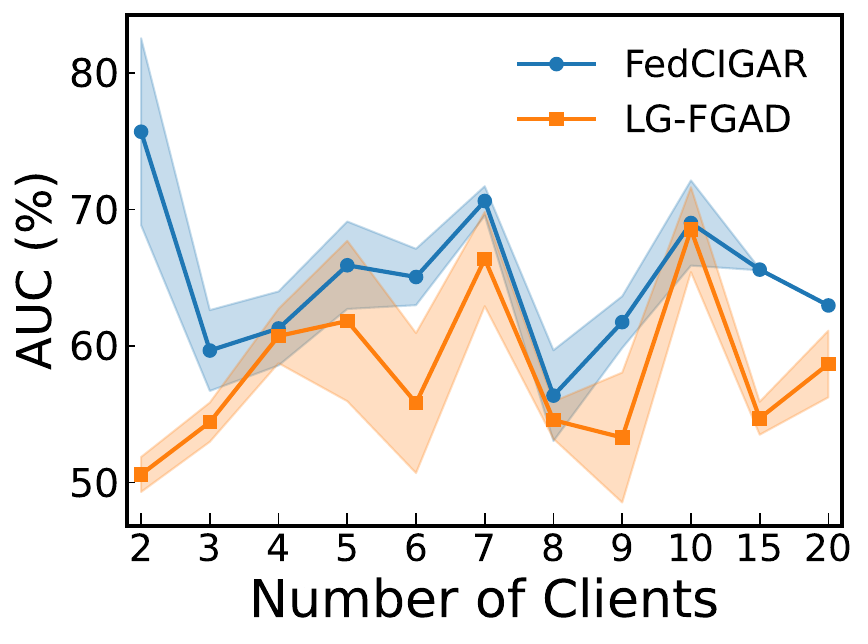}
    }\hfill
    \subfigure[MUTAG\label{subfig:ClientNum_MUTAG}]{
        \includegraphics[scale=0.27]{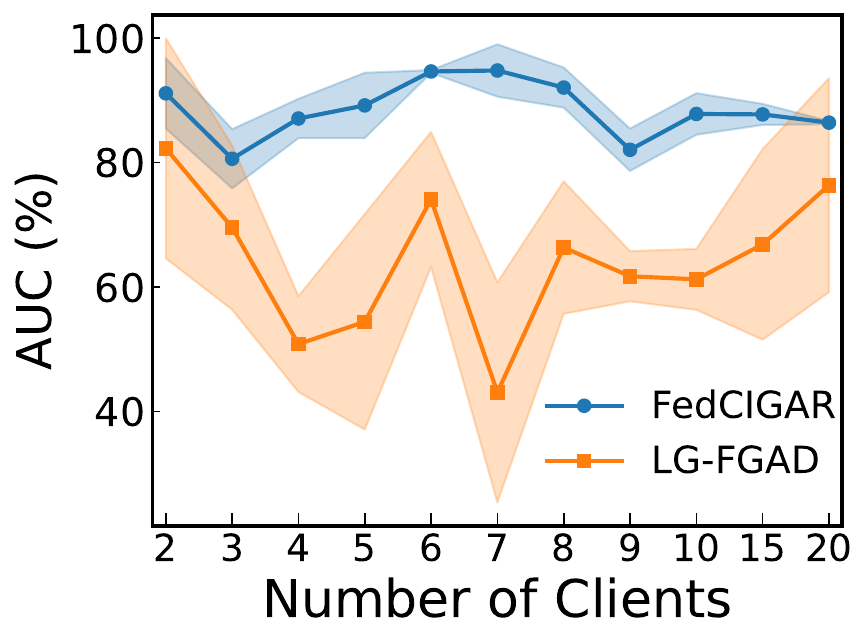}
    }
    \caption{The sensitivity of \ourmethod in terms of client number $C$.}
    \label{fig:hp_clientNum}
\end{figure}
\noindent\textbf{Client Number $C$.} To evaluate the performance of \ourmethod across different FedGLAD scenarios, we split IMDB-BINARY and MUTAG datasets into 2-20 shards, with each shard assigned to a client. As shown in Fig.~\ref{fig:hp_clientNum}, increasing the number of clients causes fluctuations in the performance of both algorithms, while \ourmethod consistently outperforms the state-of-the-art baseline LG-FGAD, demonstrating its robustness. More results can be found in Appendix D.1-D.2.  

\section{Related Work}

\noindent\textbf{Graph-level anomaly detection (GLAD)} aims to detect the abnormal graphs that are different from the majority~\cite{zhang2022dual,niu2023graph,sutherland2003spline}. Since anomaly labels are difficult to obtain in many practical scenarios~\cite{li2026towards,pan2026explainable}, mainstream GLAD studies primarily focus on a one-class learning setting, i.e., models are trained using only normal instances while anomalies are detected as deviations at test time. 
In recent years, graph neural networks (GNNs) have been widely applied across various tasks\cite{li2026ofa,miao2026blindguard,zheng2022unsupervised,qian2026dynhd,tan2025bisecle} and have become the dominant paradigm for GLAD, with representative work exploring reconstruction-based approaches~\cite{kim2024rethinking}, inter-graph relational mining~\cite{ma2023towards} and interpretable frameworks~\cite{liu2023towards} (see Appendix E.1 for detailed review). Despite their strong performance, these GNN-based methods~\cite{liu2026few,pan2026correcting,pan2025survey,zhao2025freegad,chen2025uncertainty} rely on centralized training, where all graph data are collected at a central server. However, in real-world scenarios, 
graph data are often distributed across multiple clients with non-IID characteristics, highlighting the significance of developing collaborative and privacy-preserving GLAD approaches in a federated setting. 

\noindent\textbf{Federated graph learning (FGL)} extends federated learning (FL) to graph-structured data with distributed and privacy-preserving training~\cite{liu2025federated,guo2023globally,wang2022graphfl}. Owing to the diversity of graph data, the data heterogeneity problem (i.e., non-IID data distributions across different clients) is an important challenge in FGL. Prior works have approached this problem through different aspects, such as client clustering~\cite{xie2021federated} and structure-aware knowledge sharing~\cite{tan2023federated} (see Appendix E.2 for detailed review). Despite their remarkable success, most existing studies focus on node or graph classification, while few studies explore FGL for GLAD. Among these, FGAD~\cite{cai2024towards} and LG-FGAD~\cite{cai2024lg} employ knowledge distillation but rely on the FedAVG-style collaborative learning strategy, which may fail on heterogeneous datasets. To address these issues, we introduce a sliding clustering and anomaly gate module to design an FGL algorithm for GLAD.

\section{Conclusion}

In this paper, we propose \ourmethod, a novel reconstruction-based framework for federated graph-level anomaly detection (FedGLAD), which addresses the detection challenges under heterogeneous graph data. \ourmethod uses a feature-structure joint reconstruction mechanism to capture anomaly patterns without annotated labels. It also integrates a contribution gating module with a sliding window-based clustering aggregation strategy to effectively alleviate data heterogeneity. Extensive experiments on real-world datasets from various domains demonstrate that \ourmethod consistently outperforms state-of-the-art methods in terms of detection performance.

\clearpage
\section*{Acknowledgments}

This work was partially supported by the Specific Research Project of Guangxi for Research
Bases and Talents (GuiKe AD24010011), the Key Research Development Program Project
of Guangxi (GuiKe AB25069095).

\bibliographystyle{named}
\bibliography{ijcai25}

\clearpage

\end{document}